\documentclass[10pt,twocolumn,letterpaper]{article}

\usepackage[pagenumbers]{cvpr} % To force page numbers, e.g. for an arXiv version

\usepackage{times}
\usepackage{epsfig}
\usepackage{graphicx}
\usepackage{amsmath}
\usepackage{amssymb}

\usepackage{algorithm}
\usepackage{algorithmic}
\usepackage{multirow}
\usepackage{makecell}
\usepackage{booktabs}
\usepackage{siunitx}
\usepackage{float}
\usepackage{verbatim}
\usepackage{color}
\usepackage{bm}
\usepackage{xspace}
\usepackage{graphicx}
\usepackage{subcaption}
\usepackage{appendix}

\def \alambic {\includegraphics[width=0.03\linewidth]{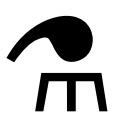}\xspace}

\usepackage[pagebackref,breaklinks,colorlinks]{hyperref}

\usepackage[capitalize]{cleveref}
\crefname{section}{Sec.}{Secs.}
\Crefname{section}{Section}{Sections}
\Crefname{table}{Table}{Tables}
\crefname{table}{Tab.}{Tabs.}

\begin{document}

%%%%%%%%% TITLE
\title{DearKD: Data-Efficient Early Knowledge Distillation for Vision Transformers}

\author{Xianing Chen$^{1}$\footnotemark[1] \ , Qiong Cao$^{2}$\footnotemark[2] \ , Yujie Zhong$^3$, Jing Zhang$^4$, Shenghua Gao$^{1,5,6}$\footnotemark[2]\ , Dacheng Tao$^{2,4}$ \\
$^1$ShanghaiTech University, 
$^2$JD Explore Academy, 
$^3$Meituan Inc., \\
$^4$The University of Sydney, 
$^5$Shanghai Engineering Research Center of Intelligent Vision and Imaging, \\
$^6$Shanghai Engineering Research Center of Energy Efficient and Custom AI IC \\
\tt\small{\{chenxn1,gaoshh\}@shanghaitech.edu.cn \quad \{mathqiong2012,dacheng.tao\}@gmail.com} \\
\tt\small{jaszhong@hotmail.com \quad jing.zhang1@sydney.edu.au} \\
}

\maketitle
\footnotetext[1]{This work was done when Xianing Chen was intern at JD Explore
Academy.}
\footnotetext[2]{Corresponding authors.}

\begin{abstract}
Transformers are successfully applied to computer vision due to their powerful modeling capacity with self-attention. However, the excellent performance of transformers heavily depends on enormous training images. Thus, a data-efficient transformer solution is urgently needed. 
In this work, we propose an early knowledge distillation framework, which is termed as DearKD, to improve the data efficiency required by transformers. Our DearKD is a two-stage framework that first distills the inductive biases from the early intermediate layers of a CNN and then gives the transformer full play by training without distillation. Further, our DearKD can be readily applied to the extreme data-free case where no real images are available. In this case, we propose a boundary-preserving intra-divergence loss based on DeepInversion to further close the performance gap against the full-data counterpart.
Extensive experiments on ImageNet, partial ImageNet, data-free setting and other downstream tasks prove the superiority of DearKD over its baselines and state-of-the-art methods. 
\end{abstract}

%%%%%%%%% BODY TEXT
\section{Introduction}
Transformers \cite{vaswani2017attention, devlin2018bert, brown2020language} have shown a domination trend in NLP studies owing to their strong ability in modeling long-range dependencies by the self-attention mechanism. Recently, transformers are applied to various computer vision tasks and achieve strong performance~\cite{dosovitskiy2020image, chen2021pre, liu2021swin}. 
However, transformers require an enormous amount of training data since they lack certain inductive biases (IB) \cite{dosovitskiy2020image, d2021convit, touvron2021training, xu2021vitae}. 
Inductive biases can highly influence the generalization of learning algorithms, independent of data, by pushing learning algorithms towards particular solutions \cite{mitchell1980need, gordon1995evaluation, goyal2020inductive}.
Unlike transformers, CNNs are naturally equipped with strong inductive biases by two constraints: locality and weight sharing mechanisms in the convolution operation. 
Thus, CNNs are sample-efficient and parameter-efficient due to the translation equivariance properties \cite{simoncelli2001natural, ruderman1994statistics, d2021convit}.

\begin{figure}[t]
	\centering
	\includegraphics[width=240pt, height=143pt]{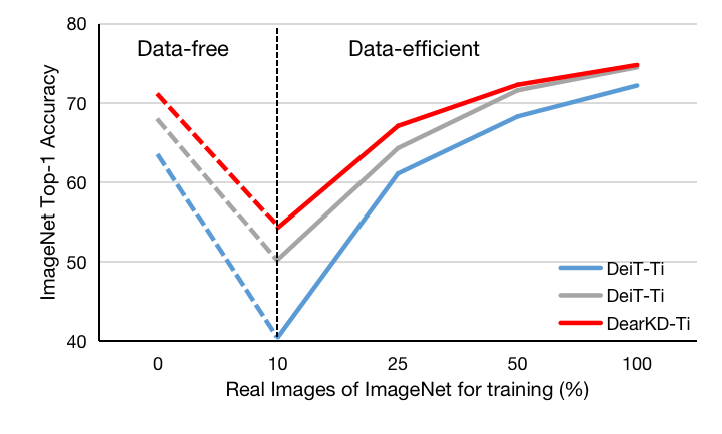}
	\caption{\textbf{Illustration of data-efficient of our DearKD.} We compare the data-efficient properties of DearKD in three situations with different numbers of real training images: the full ImageNet, the partial ImageNet and the data-free case (i.e. without any real images) with DeiT and DeiT$\alambic$.}
	\label{effi}
\end{figure}

Recently, some researchers have proposed to explicitly insert convolution operations into vision transformers to introduce inductive biases \cite{li2021contextual, dai2021coatnet, xiao2021early, graham2021levit, wu2021cvt, xu2021vitae, zhang2022vitaev2}. However, the forcefully modified structure may destroy the intrinsic properties in transformers and reduce their capacity.

Another line of work~\cite{touvron2021training} utilizes Knowledge Distillation (KD) \cite{hinton2015distilling} to realize data-efficient transformers.
By distillation, the inductive biases reflected in the dark knowledge from the teacher network can be transferred to the student \cite{abnar2020transferring}.
DeiT \cite{touvron2021training}, as a typical method in this line, has successfully explored the idea of distilling knowledge from CNNs to transformers and greatly increased the data efficiency of transformer training. 
Nevertheless, DeiT still suffers two drawbacks:

Firstly, some works \cite{dai2021coatnet, xiao2021early} \emph{reveal that inserting convolutions to the early stage of the network brings the best performance}, while DeiT only distills from the classification logits of the CNN and thus makes it difficult for the early (i.e. shallow) transformer layers to capture the inductive biases. 
Furthermore, the distillation throughout the training implicitly hinders transformers from learning their own inductive biases  \cite{d2021convit} and stronger representations \cite{dai2021coatnet}.  

To solve these problems, we propose a two-stage learning framework, named as Data-efficient EARly Knowledge Distillation (DearKD), to further push the limit of data efficiency of training vision transformers. 
Here the term `early' refers to two novel designs in our proposed framework: knowledge distillation in the early layers in transformers and in the early stage of transformer training.
\textbf{First}, we propose to distill from both the classification logits and the intermediate layers of the CNN, which can provide more explicit learning signals for the intermediate transformer layers (especially the early layers) to capture the inductive biases. 
Specifically, we draw the inspiration from ~\cite{2020On} and design a Multi-Head Convolutional-Attention (MHCA) layer to better mimic a convolutional layer without constraining the expressive capacity of self-attention. Further, we propose an aligner module to solve the problem of feature misalignment between CNN features and transformers tokens.
\textbf{Second}, the distillation only happens in the first stage of DearKD training. We let transformers learn their own inductive biases in the second stage, in order to fully leverage the flexibility and strong expressive power of self-attention. 

To fully explore the power of DearKD with respect to data efficiency, we investigate DearKD in three situations with different number of real training images (Figure \ref{effi}): the full ImageNet~\cite{deng2009imagenet}, the partial ImageNet and the data-free case (i.e. without any real images).
In the extreme case where no real images are available, networks can be trained using data-free knowledge distillation methods~\cite{chen2019data, micaelli2019zero,yin2020dreaming}.
In this work, we further enhance the performance of transformer networks under the data-free setting by introducing a boundary-preserving intra-divergence loss based on DeepInversion~\cite{yin2020dreaming}. The proposed loss significantly increases the diversity of the generated images by keeping the positive samples away from others in the latent space while maintaining the class boundaries.

Our main contributions are summarized as follows: 
\begin{itemize} 
\item We introduce DearKD, a two-stage learning framework for training vision transformers in a data-efficient manner. In particular, we propose to distill the knowledge of intermediate layers from CNNs to transformers in the early phase, which has never been explored in previous works.
\item We investigate DearKD in three different settings and propose an intra-divergence loss based on DeepInversion to greatly diversify the generated images and further improve the transformer network in the data-free situation. 
\item With the full ImageNet, our DearKD achieves state-of-the-art performance on image classification with similar or less computation.
Impressively, training DearKD with only 50\% ImageNet data can outperform the baseline transformer trained with all data. Last but not least, the data-free DearKD based on DeiT-Ti achieves 71.2\% on ImageNet, which is only 1.0\% lower than its full-ImageNet counterpart.
\end{itemize}

%------------------------------------------------------------------------
\section{Related work}
\textbf{Knowledge Distillation.} Knowledge Distillation \cite{hinton2015distilling} is a fundamental training technique, where a student model is optimized under the effective information transfer and supervision of a teacher model or ensembles. Hinton \cite{hinton2015distilling} performed knowledge distillation via minimizing the distance between the output distribution statistics between student and teacher networks to let the student learn dark knowledge that contains the similarities between different classes, which are not provided by the ground-truth labels. To learn knowledge from teacher network with high fidelity, \cite{zagoruyko2016paying} further took advantage of the concepts of attention to enhance the performance of the student network. \cite{heo2019knowledge} focus on transferring activation boundaries formed by hidden neurons. \cite{srinivas2018knowledge} proposed to match the Jacobians. \cite{liu2019structured} proposed to distill the structured knowledge. Moreover,  \cite{jiao2019tinybert} proposed a Transformers distillation method to transfer  the plenty of knowledge encoded in a large  BERT\cite{devlin2018bert} to a small student Transformer network. However, all of them do not consider the problem of distillation between two networks with different architectures. Moreover, the teacher network has lower capacity than the student network in our setting.

\textbf{Vision Transformers.} With the success of Transformers \cite{vaswani2017attention} in natural language processing, many studies \cite{dosovitskiy2020image,  chen2021pre, touvron2021training, ramachandran2019stand} have shown that they can be applied to the field of computer vision as well. Since they lack inductive bias, they indeed learn inductive biases from amounts of data implicitly and lag behind CNNs in the low data regime \cite{dosovitskiy2020image}. Recently, some works try to introduce CNNs into vision transformers explicitly \cite{chen2021oh, li2021contextual, dai2021coatnet, xiao2021early, graham2021levit, wu2021cvt, xu2021vitae}. However, their forcefully modified structure destroyed the intrinsic properties in transformers. \cite{d2021convit} introduced local inductive bias in modeling local visual structures implicitly, which still learns local information through training from amounts of data. \cite{touvron2021training} proposed to distill knowledge from CNNs to transformers which does not consider the differences in their inherent representations and the Transformers intrinsic inductive biases. Thus, we propose the two-stage learning framework for Transformers to learn convolutional as well as their own inductive biases.

\textbf{Data-Free KD.} Data-Free KD \cite{lopes2017data} aims to learn a student model from a cumbersome teacher without accessing real-world data. The existing works can be roughly divide into two categories: GAN-based and prior-based methods. GAN-based methods \cite{micaelli2019zero, ye2020data, chen2019data, zhang2021data} synthesized training samples through maximizing response on the discriminator. Prior-based methods \cite{chawla2021data} provide another perspective for data-free KD, where the synthetic data are forced to satisfy a pre-defined prior, such as total variance prior \cite{Mordvintsev2015InceptionismGD, bhardwaj2019dream} and batch normalization statistics \cite{chen2019data, chawla2021data}. However, they all has the problem of mode collapse\cite{srivastava2017veegan, che2016mode}, so we propose a boundary-preserving intra-divergence loss for DeepInversion \cite{yin2020dreaming} to generate diverse samples.

\begin{figure}[t]
	\centering
	\includegraphics[width=240pt, height=300pt]{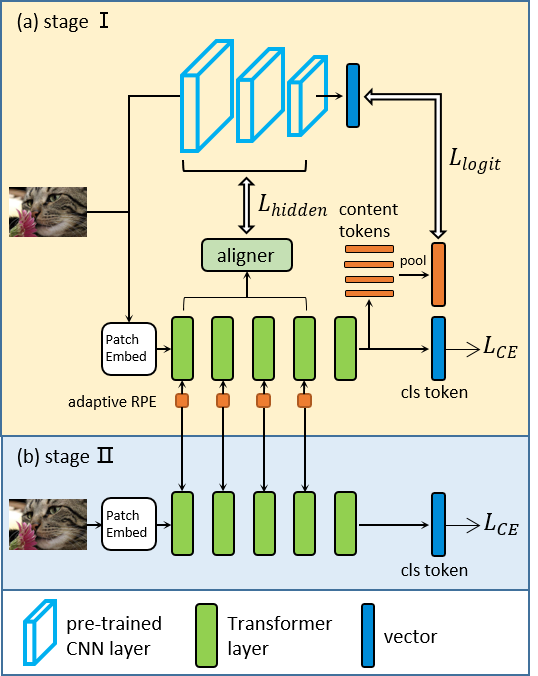}
	\caption{\textbf{The pipeline of our proposed method.} (a) The convolutional inductive biases knowledge distillation phase. (b) The transformers instrinsic inductive biases learning phase.}
	\label{pipeline}
\end{figure}

%------------------------------------------------------------------------
\section{Data-efficient Early Knowledge Distillation}
In this section, we first recap the preliminaries of Vision Transformers, and then introduce our proposed two-stage learning framework DearKD. 

\textbf{Preliminary.} Vanilla multi-head self-attention (MHSA) \cite{vaswani2017attention} is based on a trainable associative memory with (key, value) vector pairs. Specifically, input sequences $X\in R^{T \times d}$ are first linearly projected to queries (Q), keys (K) and values (V) using projection matrices, i.e. $\left( Q, K, V \right)= \left( XW^{Q}, XW^{K}, XW^{V} \right)$, where $W^{Q/K/V} \in R^{d \times d}$ denotes the projection matrix for query, key, and value, respectively. Then, to extract the semantic dependencies between each parts, a dot product attention scaled and normalized with a Softmax layer is performed. The sequences of values are then weighted by the attention. This self-attention operation is repeated $h$ times to formulate the MHSA module, where $h$ is the number of heads. Finally, the output features of the h heads are concatenated along the channel dimension to produce the output of MHSA.

\begin{equation}
\begin{array}{l}
\text{MHSA}(\boldsymbol{X}) = \boldsymbol{A}\boldsymbol{X}\boldsymbol{W}^{V} \\
\boldsymbol{A} = \text{Softmax}(\boldsymbol{Q}\boldsymbol{K})
\end{array}
\label{eq1}
\end{equation}

\textbf{Inductive Biases Knowledge Distillation.} 
It is revealed in~\cite{dai2021coatnet, xiao2021early} that convolutions in the early stage of the network can significantly enhance the performance since local patterns (like texture) can be well captured by the convolution in the early layers.
Therefore, providing explicit guidance of inductive biases to the early transformer layers becomes crucial for improving data efficiency.
However, in the later phase, this guidance may restrict the transformer from fully exploring its expressive capacity. 
To this end, we propose a two-stage knowledge distillation framework DearKD (Figure~\ref{pipeline}) for learning inductive biases for transformers, which is elaborated in the following.

\subsection{DearKD: Stage \uppercase\expandafter{\romannumeral1}}

\textbf{Multi-Head Convolutional-Attention (MHCA).} 
Recently, \cite{2020On} proves that a multi-head self-attention layer with $N_{h}$ heads and a relative positional encoding of dimension $D_{p} \geq 3$ can express any convolutional layer of kernel size $\sqrt{N_{h}} \times \sqrt{N_{h}}$ by setting the quadratic encoding:

\hspace{0.1pt}
\begin{equation}
\begin{array}{l}
\boldsymbol{v}^{(h)}:=-\alpha^{(h)}\left(1,-2 \boldsymbol{\Delta}_{1}^{(h)},-2 \boldsymbol{\Delta}_{2}^{(h)}\right) \quad \\ \boldsymbol{r}_{\delta}:=\left(\|\boldsymbol{\delta}\|^{2}, \boldsymbol{\delta}_{1}, \boldsymbol{\delta}_{2}\right) \quad \\ 
\boldsymbol{W}_{\text{qry}}=\boldsymbol{W}_{\text{key}}:=\mathbf{0}, \quad \widehat{\boldsymbol{W}_{\text {key }}}:=\boldsymbol{I} 
\end{array}
\label{eq2}
\end{equation}

\noindent where the learned parameters $\boldsymbol{\Delta}^{(h)}=\left(\boldsymbol{\Delta}_{1}^{(h)}, \boldsymbol{\Delta}_{2}^{(h)}\right)$ and $\alpha^{(h)}$ control the center and width of attention of each head, $\boldsymbol{\delta}=\left(\boldsymbol{\delta}_{1}, \boldsymbol{\delta}_{2}\right)$ is fixed and indicates the relative shift between query and key pixels.

Motivated by \cite{2020On}, we propose a Multi-Head Convolutional-Attention (MHCA) layer to enable a transformer layer to act as a convolution layer by using the relative positional self-attention \cite{ramachandran2019stand}. Specifically, given an input $X\in R^{T \times d} $, our MHCA layer performs multi-head self-attention as follows:

\begin{equation}
\begin{array}{l}
	\text{MHCA}(\boldsymbol{X})=\boldsymbol{A}\boldsymbol{X}\boldsymbol{W}^{V} \\
	\boldsymbol{A}=\text{Softmax}(\boldsymbol{Q}\boldsymbol{K} + \boldsymbol{v}^{(h)}\boldsymbol{r}_{ij})
\end{array}
\label{eq_mhca}
\end{equation}

\noindent where $v^{(h)}$ contains a learnable parameter $\alpha^{(h)}$ (see Equation (\ref{eq2})) to adaptively learn appropriate scale of the relative position embedding (adaptive RPE). To prevent the network from falling into the local optimum where the attention highly focuses on the local information, we add a dropout layer after the adaptive RPE. 

Different from MHSA in Equation (\ref{eq1}), the proposed MHCA consists of two parts, i.e., the content part and position part, to incorporate the relative positional information. The former learns the non-local semantic dependencies described above, and the latter makes the attention aware of local details.   

\begin{figure}[t]
	\centering
	\includegraphics[width=186pt, height=112pt]{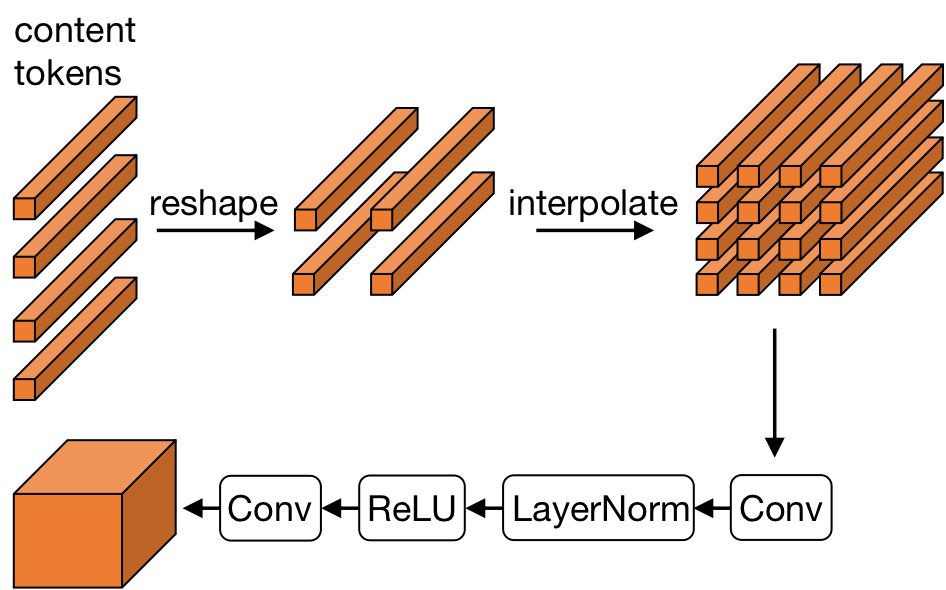}
	\caption{\textbf{Illustration of the aligner.} The aligner aligns transformer tokens to have the same size of convolution features by the stacking of reshape, bilinear interpolate, depth-wise convolution, LayerNorm and ReLU layers.}
	\label{aligner}
	\vspace{-4mm}
\end{figure}

\textbf{Early Knowledge Distillation.}
Now we consider the distillation of the convolutional inductive biases with the proposed MHCA. To capture the inductive biases and provide rich spatial information and local visual patterns for the intermediate transformer layers, we propose to distill from the intermediate layers of the CNN to transformers in the first stage. The objective is formulated as follows:    

\begin{equation}
	L_{\text{hidden}}=MSE(\text{aligner}(H^{S}), H^{T})
\end{equation}

\noindent where $H^{S} \in R^{l \times d}$ and $H^{T} \in R^{h \times w \times c}$ refer to the content tokens of student and the feature map of teacher networks respectively.
The major difficulty is that the feature maps of the CNN and the transformer tokens are in different shapes, and therefore it is infeasible to apply a distillation loss on top directly.
To tackle the problem of feature misalignment, we design an aligner module to match the size of the content tokens $H^{S}$ to that of $H^{T}$ by the stacking of reshape. 
As shown in Figure \ref{aligner}, the aligner includes a depth-wise convolution \cite{tan2019efficientnet}, LayerNorm \cite{ba2016layer} and ReLU layers. Note that, to the best of our knowledge, this work is the first to explore the knowledge distillation from the intermediate layers of the CNNs to transformers.

In addition to imitating the behaviors of intermediate CNN layers, we adopt the commonly used divergence between the teacher and student network logits in knowledge distillation. Instead of adding an additional distillation token \cite{touvron2021training} which requires additional trained CNNs networks when fine-tuning on downstream tasks, we directly pool the content tokens following \cite{heo2021rethinking, pan2021scalable} which contains discriminative information and is consistent with the design principles of CNNs. The objective with hard-label distillation \cite{touvron2021training} is as follow:

\begin{equation}
	L_{\text{logit}}=L_{\text{CE}}(logit, y_{t})
\end{equation}
\noindent where $y_{t}=argmax(logit_{T})$ is the hard decision of the teacher. 

The overall loss function is as follows:

\begin{equation}
    L = \alpha L_{\text{CE}} + (1-\alpha) L_{\text{logit}} + \beta L_{\text{hidden}}
\end{equation}
% Does Knowledge Distillation Really Work? 4.2.  
\noindent where $L_{\text{CE}}$ is the cross-entropy loss for the [CLS] token.

\subsection{DearKD: Stage \uppercase\expandafter{\romannumeral2}}
\textbf{Transformers Instrinsic Inductive Biases Learning.} Considering that transformers have a larger capacity than CNNs, we propose to encourage the transformers to learn their own inductive biases in a second stage. This is a critical step to leverage their flexibility and strong expressive
power fully. To this end, we formulate the objective of stage \uppercase\expandafter{\romannumeral2} as follows:

\vspace{-4mm}
\begin{equation}
	L = L_{\text{CE}}(logit, y)
\end{equation}

\begin{figure}[t]
	\centering
	\includegraphics[width=200pt, height=110pt]{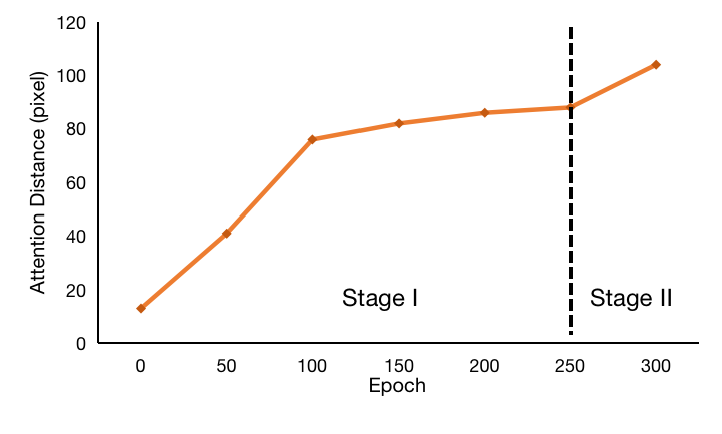}
	\caption{\textbf{The average attention distance of our DearKD for each epoch.}}
	\label{distance}
	\vspace{-4mm}
\end{figure}

\noindent Note that the relative position encoding in stage   \uppercase\expandafter{\romannumeral1} is unchanged. In this stage, the network will learn to explore a larger reception field to form the non-local representation automatically. We calculate the average attention distance of each layer in DearKD for each epoch. The results are shown in Figure \ref{distance}.  It can be observed that with the usage of convolutional IBs knowledge distillation, the transformer layers in the first stage will focuse on
modeling locality. After training our model in the second stage, the model escapes the locality, and thus, the intrinsic IBs of Transformers can be learned automatically.

\section{DF-DearKD: Training without Real Images}

\begin{figure}[t]
	\centering
	\includegraphics[width=236pt, height=148pt]{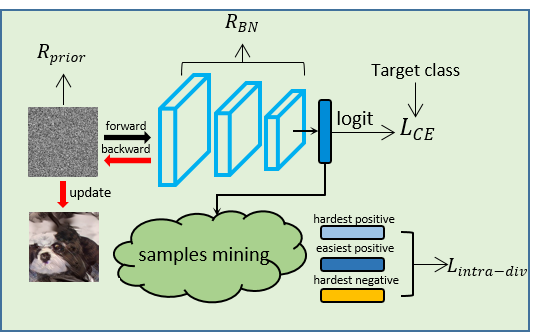}
	\caption{\textbf{The pipeline of our proposed DF-DearKD.}}
	\label{DF-DearKD}
\vspace{-4mm}
\end{figure}

To fully explore the power of DearKD with respect to data efficiency, we investigate it in the extreme setting (i.e. data-free) where no real images are available. 
In this section, we propose DF-DearKD, a data-free variant of DearKD, for crafting a transformer network without accessing any real image. 
Compared to DearKD, DF-DearKD has an extra image generation component, as illustrated in Figure \ref{DF-DearKD}. 
In the following, we first briefly review the closely related method DeepInversion~\cite{yin2020dreaming}, and then introduce a novel boundary-preserving intra-divergence loss to further increase the diversity of the generated samples.

\textbf{DeepInversion.} Assume that we have access to a trained convolution classifier as a teacher model. Given a randomly initialized input $x \in \mathbb{R}^{H \times W \times C}$ and the corresponding target label y, DeepInversion~\cite{yin2020dreaming} synthesized the image by optimizing

\vspace{-4mm}
\begin{equation}
	x = \arg\min _{x} L_{\text{CE}}(x,y)+
	R(x)+L_{\text{diversity}}(x, y)
\end{equation}

\noindent where $L_{\text{CE}}(\cdot)$ is the cross-entropy loss for classification.  $R(\cdot)$ is the image regularization term to steer $x$ away from unrealistic images and towards the distribution of images presented. $L_{\text{diversity}}(\cdot)$ is the diversity loss to avoid repeated and redundant synthetic images. Specifically, $R$ consists of two terms: the prior term $R_{prior}$ \cite{Mordvintsev2015InceptionismGD} that acts on image priors and the BN regularization term $R_{\text{BN}}$ that regularizes feature map distributions:

\begin{equation}
	R(x) = R_{\text{prior}}(x) + R_{\text{BN}}(x)
\end{equation}
\noindent where $R_{\text{prior}}$ penalizes the total variance and l2 norm of $x$, respectively. $R_{\text{BN}}$ matches the feature statistics, i.e., channel-wise mean $\mu(x)$ and variance $\sigma^{2}(x)$ of the current batch to those cached in the BN \cite{ioffe2015batch} layers at all levels.

\begin{figure}[t]
	\centering
	\subfloat[DeepInversion]{
		\centering
		\includegraphics[width=60pt, height=80pt]{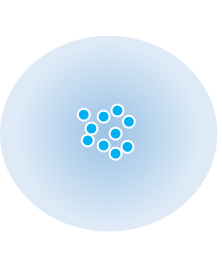}
		\label{3-a}
	}
	\hfill
	\subfloat[ADI]{
		\centering
		\includegraphics[width=60pt, height=80pt]{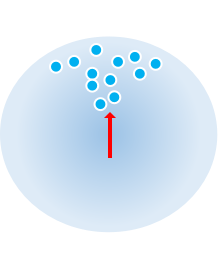}
		\label{3-b}
	}
	\hfill
	\subfloat[Ours]{
		\centering
		\includegraphics[width=80pt, height=80pt]{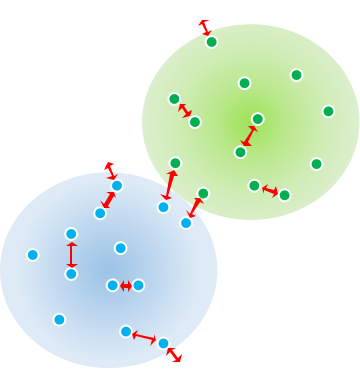}
		\label{3-c}
	}
	\hfill
	\caption{\textbf{The concept of the proposed boundary-preserving
intra-divergence loss.} Given a set of samples in the latent space (shown as dots), the boundary-preserving
intra-divergence loss in (c) pushes the easiest positive samples away from others (shown as red arrows between the same class samples) while keeping the activation boundaries (shown as circle) unaffected.}
	\label{3}
\end{figure}
% \begin{figure}[t]
% 	\centering
% 	\begin{subfigure}[b]{0.2\textwidth}
%       \includegraphics[width=60pt, height=80pt]{2-a.png}
%       \caption{DeepInversion}
%       \label{3-a}
%     \end{subfigure}
% 	\hfill
% 	\begin{subfigure}[b]{0.2\textwidth}
%       \includegraphics[width=60pt, height=80pt]{2-b.png}
%       \caption{ADI}
%       \label{3-b}
%     \end{subfigure}
% 	\hfill
% 	\begin{subfigure}[b]{0.2\textwidth}
%       \includegraphics[width=60pt, height=80pt]{2-c.png}
%       \caption{Ours}
%       \label{3-c}
%     \end{subfigure}
% 	\hfill
% 	\caption{\textbf{The concept of the proposed boundary-preserving intra-divergence loss.} Given a set of samples in the latent space (shown as dots), the boundary-preserving intra-divergence loss in (c) pushes the easiest positive samples away from others (shown as red arrows between the same class samples) while keeping the activation boundaries (shown as circle) unaffected.}
% 	\label{3}
% \end{figure}

\textbf{Boundary-preserving intra-divergence loss.}
To synthesize diverse images, Adaptive DeepInversion (ADI)\cite{yin2020dreaming} proposes a competition scheme to encourage the synthesized images out of student's learned knowledge and to cause student-teacher disagreement. However, it usually generates hard and ambiguous samples. To address the over-clustering of the embedding space (Figure \ref{3-a} and \ref{3-b}), which is similar to the mode collapse problem \cite{che2016mode, srivastava2017veegan}, we propose the boundary-preserving intra-divergence loss to keep the easiest positive samples away from others in the latent space while the class boundaries are unaffected. Figure \ref{3-c} illustrates the main idea of our proposed loss. Specifically, for each anchor image $x_{\text{a}}$ within a batch, the easiest positive samples \cite{xuan2020improved} are the most similar images that have the same label as the anchor images:

\begin{equation}
	x_{\text{ep}}=\mathop{\arg\min}_{x:C(x)=C(x_{\text{a}})} dist(f(x_{\text{a}}), f(x)) 
\end{equation}

\noindent where $dist(f(x_{a}), f(x))=\|f(x_{\text{a}})-f(x)\|_{2}$ measures the euclidean distance between two samples in the latent space. Inspired by the finding that when two latent codes are close, the corresponding images are similar \cite{xuan2020improved}, we increase the intra-class diversity by maximizing the distance between the latent code of the easiest pair of images:

\begin{equation}
	L_{\text{ep}}(x) = -dist(f(x_{\text{a}}), f(x_{\text{ep}}))
\end{equation}

This loss encourages the optimizer to explore the latent space inside the whole decision boundaries. However, this will push some generated samples out of decision boundaries. We solve this by enforcing that the anchor-positive pairs are at least closer than the anchor-negative pairs by the margin, i.e., $dist_{ap}-dist_{an} \textgreater margin$, which has the same form with the triplet loss \cite{hermans2017defense, weinberger2009distance}:

\vspace{-4mm}
\hspace{0.1pt}
\begin{equation}
	L_{\text{triplet}}(x) = max(0, dist_{\text{ap}}-dist_{\text{an}}+\text{margin})
\end{equation}

\noindent where $dist_{\text{ap}}=\|f(x_{\text{a}})-f(x_{\text{hp}})\|_{2}$ and $dist_{\text{an}}=\|f(x_{\text{a}})-f(x_{\text{hn}})\|_{2}$ measure the distance between the anchor images and the corresponding hardest positive and negative images in the latent space, respectively. And $x_{\text{hp}}=\mathop{\arg\max}_{x:C(x)=C(x_{\text{a}})} dist(f(x_{\text{a}}), f(x))$ are the hardest positive samples which are the least similar images that have the same label with the anchor images,  $x_{hn}=\mathop{\arg\max}_{x:C(x)=C(x_{\text{a}})} dist(f(x_{\text{a}}), f(x))$ are the hardest negative samples which are the most similar images which have different labels from the anchor images.
Therefore, the overall proposed intra-divergence loss is:
\hspace{0.1pt}
\begin{equation}
	L_{\text{intra-div}}(x) = \alpha_{\text{ep}} L_{\text{ep}}(x) +
	\alpha_{\text{triplet}} L_{\text{triplet}}(x)
\end{equation}

%------------------------------------------------------------------------
\section{Experiments}
In this section, we evaluate the effectiveness of our proposed DearKD on ImageNet to show that our two-stage learning framework for Transformers can boost the performance of Transformers. First, we provide an ablation study for the impact of each choice and analyze of data efficiency for transformers. Then, we compare with state-of-the-arts and investigate its generalization ability on downstream tasks. Finally, we analyse the  results of DF-DearKD.

\vspace{2mm}
\subsection{Implementation Details} 

We based our model on the DeiT \cite{touvron2021training}, which is a  hyperparameter-optimized version of ViT. Our models have three variants named DearKD-Ti, DearKD-S, DearKD-B, which are the same with DeiT-Ti, DeiT-S, DeiT-B, except that we increase the heads number of our three variants to 12, 12, 16 while keeping the vector dimension unchanged to increase the ability to represent convolution \cite{2020On, d2021convit}. Specifically, we first embed input images of size 224 into 16 $\times$ 16 non-overlapping patches. Then we propagate the patches through 8 MHCA and 4 MHSA blocks. Since the relative position embedding in MHCA is not suitable for the [CLS] token, which should disregard the positions of all other tokens, we simply pad the relative position embedding with zero vector and add them to all tokens. During testing or fine-tuning, we only use the [CLS] token to obtain the probability distribution. 
Note that our method can be easily extended to any vision transformer model.

Following \cite{touvron2021training}, we use a pre-trained RegNetY-16GF from timm \cite{rw2019timm} that achieves 82.9\% top-1 accuracy as our teacher model. Our models are trained from scratch using AdamW optimizer for 300 epochs with cosine learning rate decay. We optimize the model in the first stage with 250 epochs. The learning rate is 0.0005. When we train models with more epochs, we append the epochs number at the end, e.g. DearKD-Ti-1000, and train the model in the first stage with 800 epochs. A batch size of 2048 is used. The image size during training is set to $224 \times 224$. We use Mixup \cite{2017mixup}, Cutmix \cite{0CutMix}, Random Erasing \cite{2017Random} and Random Augmentation \cite{2017Random} for data augmentation. Experiments are conducted on 8 NVIDIA A100 GPUs.

\begin{table}[htbp]
	\centering
	\begin{tabular}{cccc|c}
		\hline
		MHCA & $L_{\text{hidden}}$ & distill & two-stage & Top1 \\
		\hline
		 & & & & 72.3 \\
		\checkmark & & & & 72.5  \\
		 & & \checkmark & & 74.3 \\
		 & \checkmark & \checkmark & & 74.1 \\
		\checkmark & \checkmark & \checkmark & & 74.6 \\
		\checkmark & \checkmark & \checkmark & \checkmark & 74.8 \\
		\hline
	\end{tabular}
	\caption{\textbf{Ablation of different modules} evaluated on ImageNet classification. DeiT-Ti and DearKD-Ti are used. Here, ·distill’ indicates the first stage of our learning framework. The symbol \checkmark indicates that we use the corresponding element.}
 	\label{TAB_abla}
 	\vspace{2mm}
\end{table}

\begin{table}[htbp]
	\centering
	\begin{tabular}{c|cc|cc|c}
		\hline
		\multirow{2}{*}{Train size} & 
		\multicolumn{2}{c|}{DeiT-Ti} & 
		\multicolumn{2}{c|}{DeiT-Ti$\alambic$} & 
		\multirow{2}{*}{DearKD-Ti} \\
		\cline{2-5}
            & Top1 & Gap & Top1 & Gap \\
		\hline
		10\% & 40.5 & 13.8\% & 50.3 & 4.0\% & 54.3 \\ %48\59.6
		25\% & 61.1  & 6.0\% & 64.3 & 2.8\% & 67.1 \\ %66.1\73.7
		50\% & 68.3 & 4.0\% & 71.6 & 0.7\% & 72.3 \\ %74.6\78.2
		100\% & 72.2 & 2.6\% & 74.5 & 0.3\% & 74.8 \\ 
		\hline
	\end{tabular}
	\caption{\textbf{Comparison of data efficiency of DearKD and DeiT on ImageNet.}}
 	\label{TAB_data}
\end{table}

\vspace{2mm}
\subsection{Ablation Study}
In this section, we ablate the important elements of our design in the proposed DearKD. We use DeiT-Ti with attention heads changed as our baseline model in the following ablation study. All the models are trained for 300 epochs on ImageNet and follow the same training setting and data augmentation strategies as described above.

As can be seen in Table \ref{TAB_abla}, using our two-stage learning framework achieves the best 74.8\% Top-1 accuracy among other settings. By adding our MHCA, our model reaches a Top-1 of 72.5\%, outperforming the original DeiT-Ti with comparable parameters. This mild improvement is mainly because of the introduction of the locality. Note that our DearKD uses pooled content tokens as our distillation token and achieves comparable performance with DeiT-Ti$\alambic$, which adds additional distillation tokens. Thus our model can be applied to downstream tasks without a pre-trained teacher model while the inductive biases are stored in the adaptive RPE in our MHCA. Since the differences between the feature representations of CNNs and Transformers, adding the hidden stage distillation loss decreases the model performance. Thanks to our proposed MHCA, the hidden stage distillation loss with our MHCA together brings +2.3\%, illustrating their complementarity. Finally, after using a two-stage learning framework which introduces the intrinsic IBs of Transformers, the performance increases to 74.8\% Top-1 accuracy, demonstrating the effectiveness of learning Transformers intrinsic IB. 

\vspace{3mm}
\subsection{Analysis of Data Efficiency}
To validate the effectiveness of the introduced inductive biases learning framework in improving data efficiency and training efficiency, we compare our DearKD with DeiT, DeiT by training them using 10\%, 25\%, 50\%, and 100\% ImageNet training set. The results are shown in Table \ref{TAB_data}.
As can be seen, DearKD consistently outperforms the DeiT baseline and DeiT\alambic by a large margin.
Impressively, DearKD using only 50\% training data achieves better performance with DeiT baseline using all data. When all training data are used, DearKD significantly outperforms DeiT baseline using all data by about an absolute 2.6\% accuracy. It is also noteworthy that as the data volume is decreased, the gap between our DearKD and DeiT is increased, which demonstrates that our method can facilitate the training of vision transformers in the low data regime and make it possible to learn more efficiently with less training data.

\begin{table}
	\centering
	\scalebox{0.95}{
	\begin{tabular}{c|c|c|c|c}
		\hline
		Method & Params & size & throughput & Top1  \\
		\hline
		\multicolumn{5}{c}{CNNs} \\
		\hline
		ResNet-18\cite{he2016deep}  & 12M & $224^{2}$ & 4458.4 & 69.8 \\
		ResNet-50\cite{he2016deep} & 25M & $224^{2}$ & 1226.1 & 76.2  \\
		ResNet-101\cite{he2016deep} & 45M & $224^{2}$  & 753.6 & 77.4  \\
		ResNet-152\cite{he2016deep} & 60M & $224^{2}$  & 526.4 & 78.3  \\
		\hline
		RegNetY-4GF\cite{radosavovic2020designing} & 21M & $224^{2}$  & 1156.7 & 80.0  \\
		RegNetY-8GF\cite{radosavovic2020designing} & 39M & $224^{2}$  & 591.6 & 81.7  \\
		RegNetY-16GF\cite{radosavovic2020designing} & 84M & $224^{2}$ & 334.7 & 82.9 \\
		\hline
		EffiNet-B0\cite{tan2019efficientnet} & 5M & $224^{2}$  & 2694.3 & 77.1  \\
% 		EffiNet-B1\cite{tan2019efficientnet}  & 8M & $240^{2}$  & 1662.5 & 79.1 \\
% 		EffiNet-B2\cite{tan2019efficientnet}  & 9M & $260^{2}$  & 1255.7 & 80.1  \\
		EffiNet-B3\cite{tan2019efficientnet}  & 12M & $300^{2}$  & 732.1 & 81.6 \\
		EffiNet-B4\cite{tan2019efficientnet}  & 19M & $380^{2}$  & 349.4 & 82.9 \\
% 		EffiNet-B5\cite{tan2019efficientnet}  & 30M & $456^{2}$  & 169.1 & 83.6  \\
		EffiNet-B6\cite{tan2019efficientnet}  & 43M & $528^{2}$ & 96.9 & 84.0  \\
		EffiNet-B7\cite{tan2019efficientnet}  & 66M & $600^{2}$  & 55.1 & 84.3 \\
		\hline
		\multicolumn{5}{c}{Transformers} \\
		\hline
		ViT-B/16\cite{dosovitskiy2020image} & 86M & $384^{2}$ & 85.9 & 77.9 \\
		ViT-L/16\cite{dosovitskiy2020image} & 307M & $384^{2}$ & 27.3 & 76.5 \\
		\hline
		T2T-ViT-7\cite{2021Tokens} & 4M & $224^{2}$ & 2638.4 & 71.7\\
		T2T-ViT-14\cite{2021Tokens} & 22M & $224^{2}$ & 1443.9 & 81.5 \\
		T2T-ViT-19\cite{2021Tokens} & 39M & $224^{2}$ & 781.0 & 81.9 \\
		\hline
		DeiT-Ti\cite{touvron2021training}  & 5M & $224^{2}$ & 2536.5 & 72.2  \\
		DeiT-S\cite{touvron2021training}  & 22M & $224^{2}$ & 940.4 & 79.8 \\
		DeiT-B\cite{touvron2021training}  & 86M & $224^{2}$ & 292.3 & 81.8  \\
		% DeiT-B\cite{touvron2021training}  & 86M & $384^{2}$ & 85.9 & 83.1  \\
		\hline
		DeiT-Ti$\alambic$\cite{touvron2021training}  & 6M & $224^{2}$ & 2529.5 & 74.5  \\
		DeiT-S$\alambic$\cite{touvron2021training}  & 22M & $224^{2}$ & 936.2 & 81.2\\
		DeiT-B$\alambic$\cite{touvron2021training}  & 87M & $224^{2}$ & 290.9 & 83.4 \\
		\hline
		DeiT-Ti$\alambic$-1000\cite{touvron2021training}  & 6M & $224^{2}$ & 2529.5 & 76.6  \\
		DeiT-S$\alambic$-1000\cite{touvron2021training}  & 22M & $224^{2}$ & 936.2 & 82.6\\
		DeiT-B$\alambic$-1000\cite{touvron2021training}  & 87M & $224^{2}$ & 290.9 & 84.2 \\
		\hline
		Swin-T\cite{liu2021swin} & 29M & $224^{2}$ & 755.2 & 81.3 \\
		Swin-S\cite{liu2021swin} & 50M & $224^{2}$ & 436.9 & 83.0 \\
		Swin-B\cite{liu2021swin} & 88M & $224^{2}$ & 278.1 & 83.3 \\
		Swin-B\cite{liu2021swin} & 88M & $384^{2}$ & 84.7 & 84.2 \\
		\hline
		DearKD-Ti & 5M & $224^{2}$ & 1416.7 & 74.8 \\
		DearKD-S & 22M & $224^{2}$ & 570.1 & 81.5 \\
		DearKD-B & 86M & $224^{2}$ & 253.7 & 83.6 \\
		\hline
		DearKD-Ti-1000 & 5M & $224^{2}$ & 1416.7 & 77.0 \\
		DearKD-S-1000 & 22M & $224^{2}$ & 570.1 & 82.8 \\
		DearKD-B-1000 & 86M & $224^{2}$ & 253.7 & 84.4 \\
		\hline
	\end{tabular}}
	\caption{\textbf{Comparison of different backbones on ImageNet classification.} Throughput is measured using the GitHub repository of \cite{rw2019timm} and a V100 GPU, following \cite{touvron2021training}.}
	\label{TAB1}
	\vspace{-2mm}
\end{table}

%\subsection{Comparison with state-of-the-art} 
\subsection{Comparison with Full ImageNet} 
We compare our DearKD with both CNNs and vision Transformers with similar model sizes in Table \ref{TAB1}. As we can see from Table 3 that our DearKD achieves the best performance compared with other methods. Compared with CNNs, our DearKD-Ti achieves a 74.8\% Top-1 accuracy, which is better than ResNet-18 with more parameters. The Top-1 accuracy of the DearKD-S model is 81.5\%, which is comparable to RegNetY-8GF which has about two times of parameters than ours. Moreover, our DearKD-S achieves a better result than ResNet-152 with only a third of the parameters, showing the superiority of inductive biases learning procedure by design. Similar phenomena can also be observed when comparing DearKD with EffiNet, which requires a larger input size than ours.

In addition, we compare with multiple variants of vision transformers. We use the same structure with ViT and DeiT except that we increase the head number while keeping the channel dimension unchanged. Thanks to our carefully designed learning framework, DearKD can boost the performance of the model with ignorable additional parameters and computation cost. DearKD outperforms T2T-ViT, which adds an additional module on ViT to model local structure. Compared with Swin Transformer, DearKD with fewer parameters also achieves comparable or better performance. For example, DearKD-S achieves better performance with Swin-T but has 7M fewer parameters, demonstrating the superiority of the proposed CMHSA and learning framework.

%\subsection{Generalization on Downstream Tasks}
\paragraph{Generalization on downstream tasks.}
%In this section, we show that our models have good generalization on downstream tasks by fine-tuning them on the training sets of several fine-grained classification tasks. 
To showcase the generalization of the proposed method, we fine-tune the DearKD models on several fine-grained classification benchmarks. 
We transfer the models initialized with DearKD on full ImageNet to several benchmark tasks: CIFAR-10/100 \cite{krizhevsky2009learning},  Flowers\cite{2008Automated}, Cars\cite{20143D}, and pre-process them follow \cite{kolesnikov2020big, dosovitskiy2020image}. The results are shown in Table \ref{TAB_gene}. It can be seen that DearKD achieves SOTA performance on most of the datasets. These results demonstrate that the good generalization ability of our DearKD even without a teacher model when fine-tuning to downstream tasks.

\begin{table}
	\centering
	\begin{tabular}{c|cccc}
		\hline
		Method & Cifar10 & Cifar100 & Flowers & Cars  \\
		\hline
		ViT-B/32\cite{dosovitskiy2020image} & 97.8 & 86.3 & 85.4 & - \\
		ViT-B/16\cite{dosovitskiy2020image} & 98.1 & 87.1 & 89.5 & - \\
		ViT-L/32\cite{dosovitskiy2020image} & 97.9 & 87.1 & 86.4 & - \\
		ViT-L/16\cite{dosovitskiy2020image} & 97.9 & 86.4 & 89.7 & - \\
		T2T-ViT-14\cite{2021Tokens} & 98.3 & 88.4 & - & - \\
		EffiNet-B5\cite{tan2019efficientnet} & 98.1 & 91.1 & 98.5 & - \\
 		DeiT-B\cite{touvron2021training} & 99.1 & 90.8 & 98.4 & 92.1 \\
		DeiT-B\alambic\cite{touvron2021training} & 99.1 & 91.3 & 98.8 & 92.9 \\
		\hline
		DearKD-Ti & 97.5 & 85.7 & 95.1 & 89.0 \\
		DearKD-S & 98.4 & 89.3 & 97.4 & 91.3 \\
		DearKD-B & 99.2 & 91.1 & 98.8 & 92.7 \\
		\hline
	\end{tabular}
	\caption{\textbf{Generalization of DearKD and SOTA methods on different downstream tasks.}}
 	\label{TAB_gene}
 	\vspace{-2mm}
\end{table}

\subsection{Performance of DF-DearKD}

\textbf{Implementation details.} For the training samples generation, we use multi-resolution optimization strategy following \cite{yin2020dreaming}. We first downsample the input to resolution $112 \times 112$ and optimize for $2k$ iterations. Then, we optimize the input of resolution $224 \times 224$ for $2k$  iterations.
We use Adam optimizer and cosine learning scheduler. Learning rates for each step are 0.5 and 0.01, respectively. We set $\alpha_{\text{TV}}=1e-4, \alpha_{l_{2}}=1e-5,\alpha_{\text{BN}}=5e-2, \alpha_{\text{ep}}=50, \alpha_{\text{triplet}}=0.5 $. We set batch size to 42 and generate 6 classes each batch randomly. Image pixels are randomly initialized i.i.d. from Gaussian noise of $\mu=0$ and $\sigma=1$. 
We use RegNetY-16GF \cite{radosavovic2020designing}  from timm \cite{rw2019timm} pre-trained on ImageNet \cite{deng2009imagenet}.  Experiments are conducted on NVIDIA TITAN X GPUs.

%\vspace{2mm}
\textbf{Performance comparison.} Table \ref{TAB_DF} shows the performance of the student model obtained with different methods. As shown in the table,
our method performs significantly better than training with other data-free methods. Although our methods achieves results lower than distillation on real images with the same number, the results are close to training from scratch with original ImageNet dataset. For example, the student model trained with our method gets only 1.0\% decrease on DeiT-Ti compared with training from scratch. 

Furthermore, the ablation experiments can be seen on the last three rows in Table \ref{TAB_DF}. The third-to-last row denotes distillation with images generated from DeepInversion without diverse loss achieves
accuracy of only 62.7\%. When further training with the diversity loss of ADI, we observe 7.4\% accuracy improvement. And by applying the our intra-divergence loss brings in 8.6\% increase.

\begin{table}
	\centering
	\begin{tabular}{ccc}
		\hline
		Teacher Network & ResNet-101 & ResNet-101\\
		Teacher Accuracy & 77.37\% & 77.37\% \\ 
		Student Network & DeiT-Ti & DeiT-S \\ 
		\hline
		\multicolumn{3}{c}{Train from scratch} \\ 
		ImageNet & 72.2\% & 79.8\% \\
		\hline
		\multicolumn{3}{c}{Distill on real images} \\
		%\hline
		ImageNet  & 74.6\% ($2.4\%\uparrow$) & 81.5\%   ($1.7\%\uparrow$)\\ 
		partial ImageNet & 72.2\% ($0.0\%\downarrow$)  & 79.1\%  ($0.7\%\downarrow$) \\ %
		\hline
		\multicolumn{3}{c}{Distill on generated samples} \\
		%\hline
		DeepInversion & 62.7\% ($9.5\%\downarrow$) & 66.3 ($13.5\%\downarrow$) \\
		ADI & 70.1\% ($2.1\%\downarrow$) & 73.1 ($6.7\%\downarrow$) \\
		DF-DearKD & 71.2\% ($1.0\%\downarrow$) & 74.0 ($5.8\%\downarrow$) \\  
		\hline
	\end{tabular}
	\caption{\textbf{Knowledge distillation results from a pre-trained ResNet-101 classifier to a ViT initialized from scratch on the ImageNet dataset.} $\uparrow$ and $\downarrow$ indicate performance increase and decrease, respectively.}
	\label{TAB_DF}
	\vspace{-2mm}
\end{table}

\begin{table}
	\centering
	\begin{tabular}{c|c}
		\hline
		Method & LPIPS \\
		\hline
		real images  & 0.708 \\
		\hline
		DeepInversion & 0.657 \\
		ADI & 0.683 \\
		DF-DearKD & 0.692 \\
		\hline
	\end{tabular}
	\caption{\textbf{Diversity quantitative comparison.} We use the
LPIPS metric to measure the
 diversity of generated images. Higher LPIPS score
indicates better diversity among the generated images.}
	\label{TAB_LPIPS}
	\vspace{-4mm}
\end{table}

\textbf{Diversity comparison.} We demonstrate the diversity by comparing the LPIPS \cite{zhang2018unreasonable, lee2018diverse} of our generated images with other methods in Table \ref{TAB_LPIPS}. We compute the distance between 4000 pairs of images. We randomly sample 4 pairs of images for each class. The highest score compared with other methods shows that our method can generate diverse images. Although there is still a gap between our generated images and real images, the generated samples can be a data source to train the high-performance model.

%--------------------------------------------------
\section{Conclusion}
In this paper, we propose DearKD, an early knowledge distillation framework, to improve the data efficiency for training transformers. DearKD is comprised of two stages: in the first stage, inductive biases are distilled from the early intermediate layers of a CNN to the transformer, while the second stage allows the transformer to make full use of its capacity by training without distillation. Moreover, we enhance the performance of DearKD under the extreme data-free case by introducing a boundary-preserving intra-divergence loss based on DeepInversion to generate diverse training samples. We conduct extensive experiments on ImageNet, partial ImageNet, data-free setting  and  other downstream tasks, and demonstrate that DearKD achieves superior performance and surpasses state-of-the-art methods.

\textbf{Acknowledgements.} This work is supported by the
Major Science and Technology Innovation 2030 ”New
Generation Artificial Intelligence” key project (No.
2021ZD0111700), and National Key R\&D Program of
China (2018AAA0100704), NSFC 61932020, 62172279,
Science and Technology Commission of Shanghai Municipality
(Grant No.20ZR1436000), and ”Shuguang Program”
supported by Shanghai Education Development Foundation
and Shanghai Municipal Education Commission. Dr. Jing
Zhang is supported by ARC FL-170100117.

{\small
\bibliographystyle{ieee_fullname}
\bibliography{DearKD}
}

\vfill
%--------------------------------------------------
\clearpage

\appendix
~\\
\noindent\textbf{\Large{Appendix}}

\section{The image regularization term of DF-DearKD}
\label{supp3}
The image regularization term $R(\cdot)$ consists of two terms: the prior term $R_{prior}$ \cite{Mordvintsev2015InceptionismGD} that acts on image priors and the BN regularization term $R_{\text{BN}}$ that regularizes feature map distributions:

\begin{equation}
	R(x) = R_{\text{prior}}(x) + R_{\text{BN}}(x)
\end{equation} 

Specifically, $R_{\text{prior}}$ penalizes the total variance and l2 norm of $x$, respectively. 

\begin{equation}
	\mathcal{R}_{\text{prior}}(x) = \alpha_{TV}\mathcal{R}_{TV}(x) +
	\alpha_{l_{2}} \mathcal{R}_{l_{2}}(x)
\end{equation}

$\mathcal{R}_{\text{BN}}$ matches the feature statistics, i.e., channel-wise mean $\mu(x)$ and variance $\sigma^{2}(x)$ of the current batch to those cached in the BN \cite{ioffe2015batch} layers at all levels:

\begin{equation}
\begin{array}{r}
		\mathcal{R}_{BN}(x)=\alpha_{BN} \sum_{l=1}^{L} \left\| \mu_{l}(x) - \mu_{l}^{BN} \right\|_{2} + \\
		 \left\| \sigma_{l}^{2}(x) - \sigma_{l}^{2 BN} \right\|_{2} 
\end{array}
\end{equation}

where L is the total number of BN layers.

\section{Generated samples from DF-DearKD}
\label{supp1}
\begin{figure}[t]
	\centering
	\includegraphics[width=240pt, height=116pt]{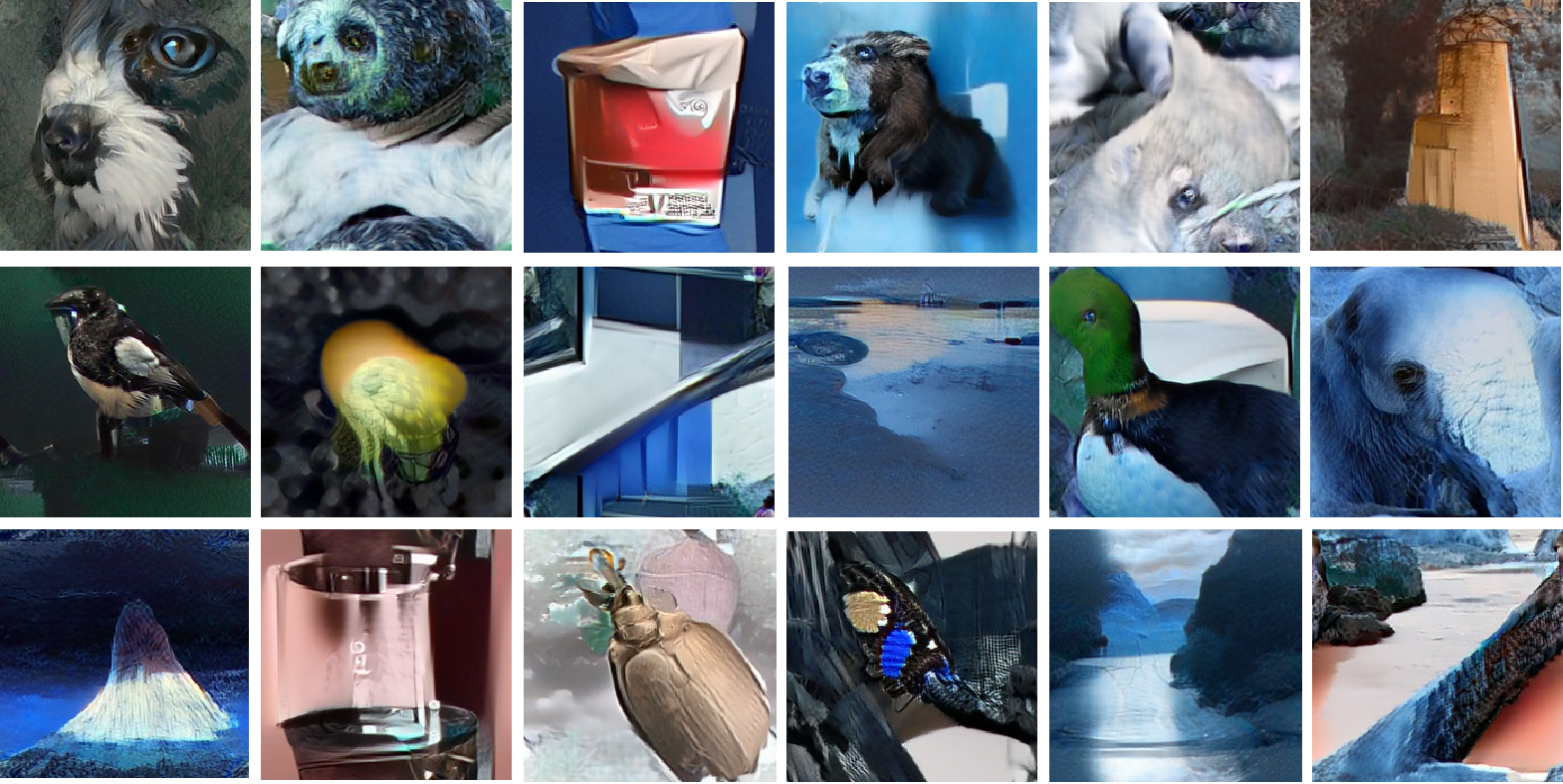}
	\caption{\textbf{Images generated by our method on RegNetY-16GF model pre-trained with ImageNet.}}
	\label{samples}
	\vspace{-4mm}
\end{figure}

Figure \ref{samples} shows samples generated by our method from an ImageNet-pretrained RegNetY-16GF model. Remarkably, given just the pre-trained teacher model, we observe that our method is able to generate images with high fidelity and resolution.

\section{Analysis of the number of epochs for the first stage of DearKD}
In this section, we ablate the number of epochs for the first stage of our DearKD. As can be seen in Table \ref{tab_epochnum}, training the model in the first stage with 250 epochs achieves the best 74.8\% Top-1 accuracy among other settings. It is not surprising that training the model in the first stage with less epochs will lead to worse performance. But, for models trained with 300 epochs, the inductive biases knowledge from CNNs are not saturated. So, we use Equation (6) in the second stage except that we set $\beta$ to 0 and let $\alpha$ linearly increase to 1. Besides, for models trained with 1000 epochs, we empirically select 800 as the number of epochs for the first stage.

\begin{table}
	\centering
	\begin{tabular}{c|ccccc}
		\hline
		Epochs number & 200 & 225 & 250 & 275 & 300 \\
		\hline
		Accuracy & 74.3 & 74.6 & 74.8 & 74.7 & 74.6 \\
		\hline
	\end{tabular}
	\caption{\textbf{Ablation of different epochs number of the first stage of DearKD evaluated on ImageNet classification.} DearKD-Ti is used.}
 	\label{tab_epochnum}
 	\vspace{-4mm}
\end{table}

\section{More implement details of DF-DearKD}
\label{supp2}
We filter out ambiguous images whose output logits from a pre-trained ResNet-101 are less than 0.1 and finally synthesize 600k images to train our transformer student network from scratch. Then, we use the target label for inversing the RegNetY-16GF as our ground truth. The RegNetY-16GF can achieve 100\% accuracy on the generated samples. This phenomenon is the same as that in \cite{yin2020dreaming}. So, we use a pre-trained ResNet-101 from pytorch \cite{paszke2019pytorch} that achieves 77.37\% top-1 accuracy on ImageNet as our teacher model, which can provide good results as well as inductive biases clues. We use AdamW optimizer with learning rate 0.0005 and cosine learning scheduler. The model is trained from scratch for 1000 epochs. A batch size of 1024 is used. We train the model in the first stage with 800 epochs.
We use Mixup \cite{2017mixup}, Cutmix \cite{0CutMix}, Random Erasing \cite{2017Random} and Random Augmentation \cite{2017Random} for data augmentation. Experiments are conducted on 4 NVIDIA TESLA V100 GPUs.

\section{Limitation and Future works}
Although DF-DearKD can generate high quality images, it still has difficulty in handling human-related classes due to the limited information stored in the feature statistics. 
Moreover, we generate lots of samples which takes a lot of time and computation costs even we do not use any real images. There is still a gap between training with generated samples and real images. In the future, we plan to investigate more in model inversion or image generation to further improve training data quality and diversity.

Besides, to further explore the data efficiency of training vision transformers under different settings (i.e. full ImageNet,  partial ImageNet and data-free case), we plan to distill other kinds of IBs for transformers and investigate how to introduce transformers' intrinsic IBs in the future study.
The data-free setting would be a particularly interesting case to cope with the emerging concern of data privacy in practice.

\end{document}